\documentclass[conference]{IEEEtran}
\usepackage{cite}
\usepackage{tikz}

\usepackage{algorithmic}
\usepackage{graphicx}
\usepackage{textcomp}
\usepackage{xcolor}
\usepackage{url}
\usepackage[utf8]{inputenc}
\usepackage[T1]{fontenc}

\DeclareUnicodeCharacter{202F}{\,}   
\DeclareUnicodeCharacter{00A0}{~}    
\DeclareUnicodeCharacter{2011}{-}    
\DeclareUnicodeCharacter{2013}{--}   
\DeclareUnicodeCharacter{2014}{---}  
\DeclareUnicodeCharacter{2019}{'}    
\DeclareUnicodeCharacter{00D7}{\times} 

\usepackage{booktabs}
\usepackage{array}

\begin{document}

\title{KrishokBondhu: A Retrieval-Augmented Voice-Based Agricultural Advisory Call Center for Bengali Farmers}

\author{
\IEEEauthorblockN{
Mohd Ruhul Ameen\textsuperscript{1}, 
Akif Islam\textsuperscript{2}, 
Farjana Aktar\textsuperscript{3}, 
M.\ Saifuzzaman Rafat\textsuperscript{4}
}
\IEEEauthorblockA{
\textsuperscript{1,2,3,4}Department of Computer Science and Engineering, University of Rajshahi\\
Rajshahi, Bangladesh\\
\textsuperscript{1}ameensunny242@ru.ac.bd, 
\textsuperscript{2}iamakifislam@gmail.com, 
\textsuperscript{3}farjana.aktar.cseru@gmail.com, 
\textsuperscript{4}saif.rafat@gmail.com
}
}

\maketitle

\begin{center}
\small Accepted at the 2026 IEEE 2nd International Conference on Quantum Photonics, Artificial Intelligence and Networking (QPAIN 2026). Copyright IEEE.
\end{center}

\begin{abstract}
In Bangladesh, many farmers continue to face challenges in accessing timely, expert-level agricultural guidance. This paper presents KrishokBondhu, a voice-enabled, call-centre–integrated advisory platform built on a Retrieval-Augmented Generation (RAG) framework, designed specifically for Bengali-speaking farmers. The system brings together authoritative agricultural handbooks, extension manuals, and NGO publications, processes them through an OCR-based pipeline to convert diverse documents into structured digital text, and indexes the curated content in a vector database to support efficient semantic retrieval. Through a simple phone-based interface, farmers can call the system to receive real-time, context-aware advice: speech-to-text converts the Bengali query, the RAG module retrieves relevant content, a large language model (Gemma 3-4B) generates a context-grounded response, and text-to-speech delivers the answer in natural spoken Bengali. In a pilot evaluation, KrishokBondhu produced high-quality responses for 72.7\% of diverse agricultural queries covering crop management, disease control, and cultivation practices. Compared to the KisanQRS benchmark, the system achieved a composite score of 4.53 (vs. 3.13) on a 5-point scale with a 44.7\% improvement with especially large gains in contextual richness (+367\%) and completeness (+100.4\%), while maintaining comparable relevance and technical specificity. Semantic-similarity analysis further revealed a strong correlation between retrieved context and answer quality, emphasizing the importance of grounding generative responses in curated documentation. KrishokBondhu demonstrates the feasibility of integrating call-centre accessibility, multilingual voice interaction, and modern RAG techniques to deliver expert-level agricultural guidance to remote Bangladeshi farmers, paving the way toward a fully AI-driven agricultural advisory ecosystem.
\end{abstract}


\section{Introduction}

Agriculture continues to play a vital role in Bangladesh’s economy. As reported in the 2022 Labour Force Survey, about 45.33\% of employment was in the agricultural sector, a rise from 40.6\% in 2016–17 \cite{labour_survey_bgd_report}. Yet, more recent estimates based on ILO‑modeled data suggest that the share has declined to around 35.27\% \cite{worldbank_agri_employment}, reflecting shifts in the labor structure. Despite its central importance, many farmers still lack reliable and timely guidance on crucial issues such as crop diseases, pest outbreaks, optimal cultivation techniques, and efficient resource use. Traditional extension services, though invaluable, are stretched too thin to provide real‑time, on‑demand support to all communities, especially in remote or underserved areas.

Another barrier is language: much of the agricultural knowledge base is written in English or technical Bengali, making it less accessible to farmers who may not be comfortable with formal or specialized terminology. At the same time, advances in large language models (LLMs) and retrieval‑augmented generation (RAG) create new opportunities to bring agricultural knowledge directly to farmers \cite{lewis2020retrieval, agask_system}. But using a general-purpose LLM without domain grounding often yields vague or incorrect advice, because it may ignore local context—such as particular crop varieties, region-specific pest cycles, or soil conditions. Systems such as AgAsk shows that when conversational systems draw directly from scientific documents through retrieval mechanisms, they are better able to provide accurate and context-aware responses in the agricultural domain \cite{agask_system}. Yet in the Bengali context, building a system that understands users’ spoken or written queries in natural language and delivers culturally appropriate guidance involves dealing with morphological complexity, dialect variation, and limited linguistic resources.

In this paper, we present KrishokBondhu, an agricultural advisory system that combines retrieval-augmented generation (RAG) with a spoken interface to offer farmers real‑time, context-aware guidance in Bengali. The name KrishokBondhu translates to “farmer’s friend,” and reflects our mission to bring expert agricultural knowledge within reach. Our system is designed to address three key challenges. First, it grounds its advice in verified agricultural sources to reduce hallucinations and vague responses. Second, it enables interaction through speech so that limited literacy does not prevent farmers from accessing guidance. Third, it adapts recommendations to Bangladesh’s specific cropping systems, climate conditions, and farming practices. The main contributions of this paper are outlined below.

\begin{enumerate}
\item We develop a Bangladesh-relevant, authoritative agricultural knowledge base and a robust OCR–cleaning–segmentation pipeline that converts heterogeneous PDFs and scanned documents into retrieval-ready text segments enriched with metadata.
\item We design and implement an end-to-end voice-enabled retrieval-augmented generation system that integrates semantic search (LanceDB), the Gemma 3–4B language model, and Bengali speech-to-text and text-to-speech modules to provide context-grounded agricultural advice over phone-based interfaces.
\item We propose a practical evaluation framework suited to low-resource settings where standardized Bengali agricultural QA benchmarks are unavailable. The framework combines expert assessment with structured comparison to published Kishan QRS examples, allowing us to quantify improvements in completeness, contextual richness, and overall response quality.
\end{enumerate}

\section{Related Work}

Agricultural advisory systems have progressed from early rule‑based and vocabulary-driven approaches toward neural and multimodal architectures. FAO’s AGROVOC, a multilingual agricultural thesaurus, has long supported indexing and cross-lingual retrieval \cite{agrovoc2022}, but its controlled‑vocabulary approach struggles with expressive, free-form farmer queries lacking flexibility or semantic inference.

More recently, researchers have explored neural question-answering approaches specifically adapted for agricultural applications. The KisanQRS system trains deep models over Kisan Call Centre logs to map farmer queries to responses \cite{kakkar2021kishan}. However, dataset access is limited, and such systems often lack grounding in authoritative texts or visual evidence. Systems like AgroLLM extend this by integrating RAG to improve the relevance and correctness of responses using agricultural databases \cite{samuel2025agrollm}.

To overcome hallucination and data sparsity, Retrieval-Augmented Generation (RAG) has become a dominant paradigm. By retrieving supporting context before generation, RAG grounds outputs in factual sources \cite{lewis2020retrieval, guu2020retrieval, izacard2021leveraging}. While RAG is already adopted in legal, medical, and code domains, its application in agriculture—especially in low-resource language contexts—remains underexplored.

Voice-based advisory approaches have been studied to overcome literacy barriers. Surveys of voice assistants in agriculture summarize their promise and challenges in rural deployment \cite{sathish2024voice_survey}. In greenhouse trials, voice messaging systems combined with human‑sensor inputs have been used to build agricultural knowledge over time \cite{uchihira2020voice}. However, most such systems rely on recording or message playback rather than interactive, context-aware conversational responses \cite{lokhande2023voicebot}

Multimodal AI is another emerging direction. AgriDoctor fuses image, text, and knowledge retrieval to build a multimodal assistant for crop disease diagnosis and domain-aware QA \cite{zhang2025agridoctor}. Similarly, spatial-vision systems using Earth Observation and retrieval-augmented methods enable conversational assessments of agricultural plots \cite{canada2025multimodal}. These systems point to the future of combining multiple modalities in agricultural advisory.

Nevertheless, gaps remain: (i) few systems fully integrate speech input, multimodal grounding, and localized domain knowledge; (ii) many are not evaluated in low-resource or local-language settings; (iii) adaptation to regional cropping systems and dialects is rare. Our work addresses these gaps by offering a voice-enabled RAG system made for Bangladeshi agriculture, combining text and voice modalities, and evaluating on domain-specific queries in Bengali.

\section{System Architecture and Methodology}

\subsection{Data Collection and Source Curation}

KrishokBondhu's knowledge base is constructed from authoritative agricultural documentation published by governmental and non-governmental organizations in Bangladesh. Key sources include the \textit{Krishi Projukti Hatboi}~\cite{barc2022handbook}, Bangladesh Agricultural Research Council (BARC) handbooks, Department of Agricultural Extension (DAE) field manuals~\cite{dae2021manual}, Bengali agricultural science textbooks, and sector-specific publications from the WorldFish Digital Repository~\cite{worldfish2022}. Key sources are summarized in Table~\ref{tab:data_sources}. The curated corpus comprises approximately 2,500 pages encompassing major crops (rice, wheat, jute, vegetables, pulses, and oilseeds), as well as livestock management, fisheries, and integrated farming practices. Source materials vary from well-structured digital PDFs to low-quality scanned images, necessitating a robust and adaptive document processing pipeline capable of handling OCR-based text extraction, error correction, and content normalization.

\begin{table}[t]
\footnotesize
\centering
\caption{Key Sources for Agricultural Knowledge Base}
\setlength{\tabcolsep}{3pt}
\begin{tabular}{p{2.6cm} p{3.1cm} p{0.9cm}}
\toprule
\textbf{Source} & \textbf{Summary} & \textbf{Pg.} \\
\midrule

Krishi Projukti Haatboi & Bengali handbook on crops and cultivation. & 608 \\

BARC Handbook & National reference on varieties, irrigation, pest control. & 221 \\

DAE Manuals & Guides for pest, soil, and irrigation management. & 225 \\

Agri Textbooks & School-level materials on crops and soil. & 300 \\

Agri Science Vol.~1 & Advanced plant physiology and crop science. & 450 \\

Farmers’ Guidebook & Aquaculture, horticulture, integrated farming. & 175 \\

BARC Bulletins & Short crop-specific technical manuals. & 50 \\

West Bengal Agro Books & Regional climate-aligned references. & 325 \\
\bottomrule
\end{tabular}
\label{tab:data_sources}
\end{table}

\subsection{Document Processing and OCR Pipeline}

\begin{figure}[!t]
\centering
\includegraphics[width=0.20\textwidth]{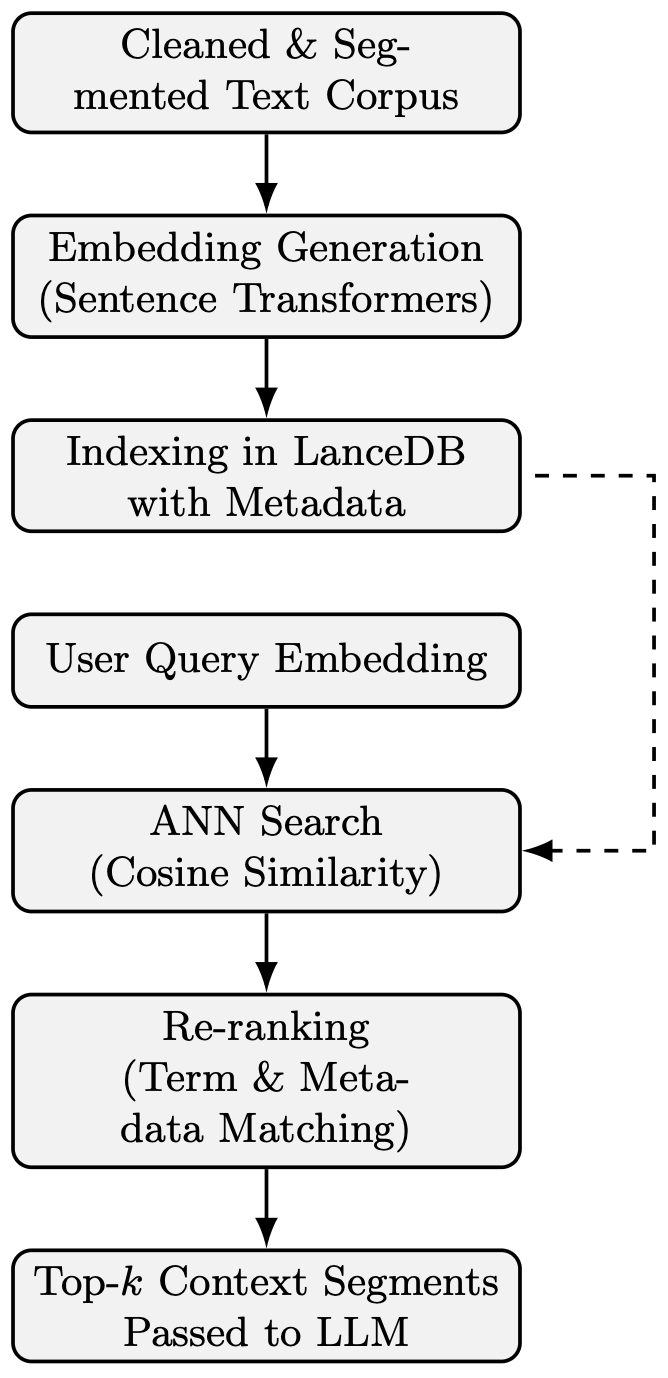}
\caption{Document Processing and OCR Pipeline}
\label{fig:ocr_pipeline}
\end{figure}

All collected documents were processed through a Bengali-capable Optical Character Recognition (OCR) and cleaning pipeline to ensure accurate text extraction and normalization. Figure~\ref{fig:ocr_pipeline} summarizes the document processing workflow from scanned pages to retrieval-ready segments. The workflow included image enhancement, skew correction, and noise reduction prior to OCR to improve recognition accuracy. Post-processing corrected common Bengali character errors using rule-based and contextual validation methods. Extracted text then underwent normalization to handle inconsistent Unicode encoding, mixed scripts, and formatting artifacts. Finally, the cleaned text was segmented into semantically coherent sections (150–300 tokens), each enriched with metadata such as source, topic, and structural position. The standardized output was stored in Markdown format to facilitate efficient vectorization and retrieval in the RAG system.

\subsection{Vector Database and Retrieval System}

\begin{figure}[h]
\centering
\includegraphics[width=0.18\textwidth]{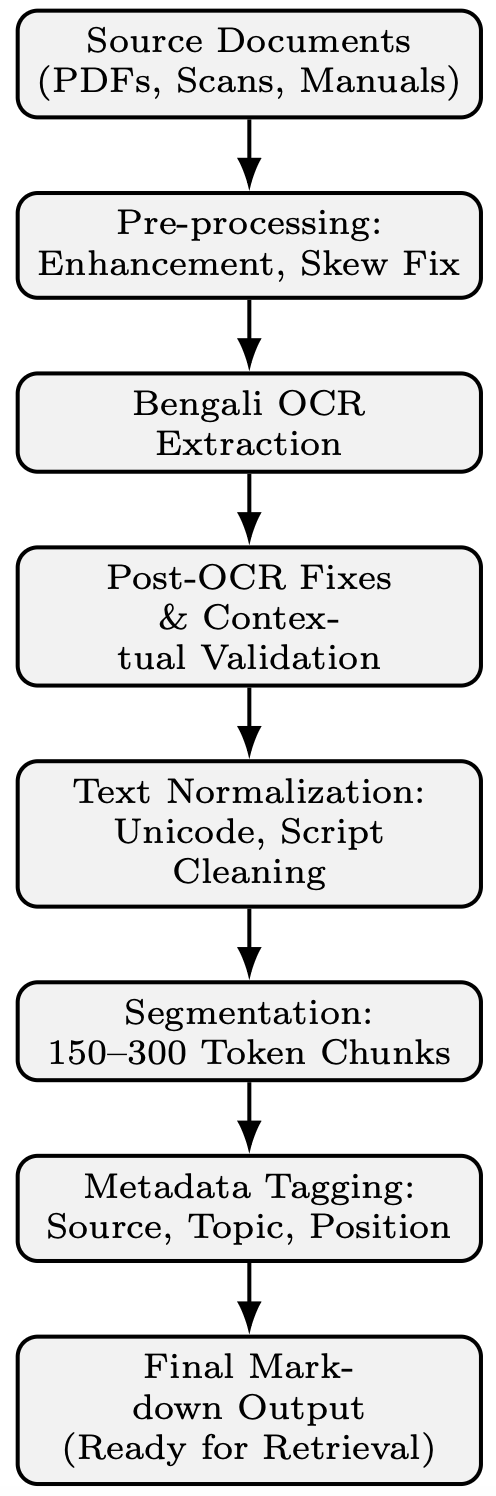}
\caption{Vector Database and Retrieval Workflow}
\label{fig:retrieval_pipeline_screenshot}
\end{figure}

Text segments were embedded using a sentence-transformer model~\cite{reimers2019sentence} to capture semantic similarity. LanceDB~\cite{lancedb2024} served as the vector backend, enabling fast approximate nearest-neighbor search with cosine similarity. At query time, user questions are embedded and matched against indexed segments, retrieving the top-$k$ results. A lightweight re-ranking layer combines semantic similarity with metadata filtering to improve precision. The retrieval workflow is illustrated in Figure~\ref{fig:retrieval_pipeline_screenshot}.

\subsection{Language Model Integration and Response Generation}

Gemma 3-4B~\cite{gemma2024}, deployed via LM Studio~\cite{lmstudio2024}, provides local response generation with strong Bengali performance. Retrieved segments are incorporated into structured prompts emphasizing factual grounding, practical guidance, and uncertainty acknowledgment. Conservative sampling parameters prioritize consistency. Post-processing checks coherence and alignment with retrieved context before formatting responses for voice delivery.

\subsection{Voice Interface Integration with VAPI}

\begin{figure}[!b]
\centering
\includegraphics[width=0.35\textwidth]{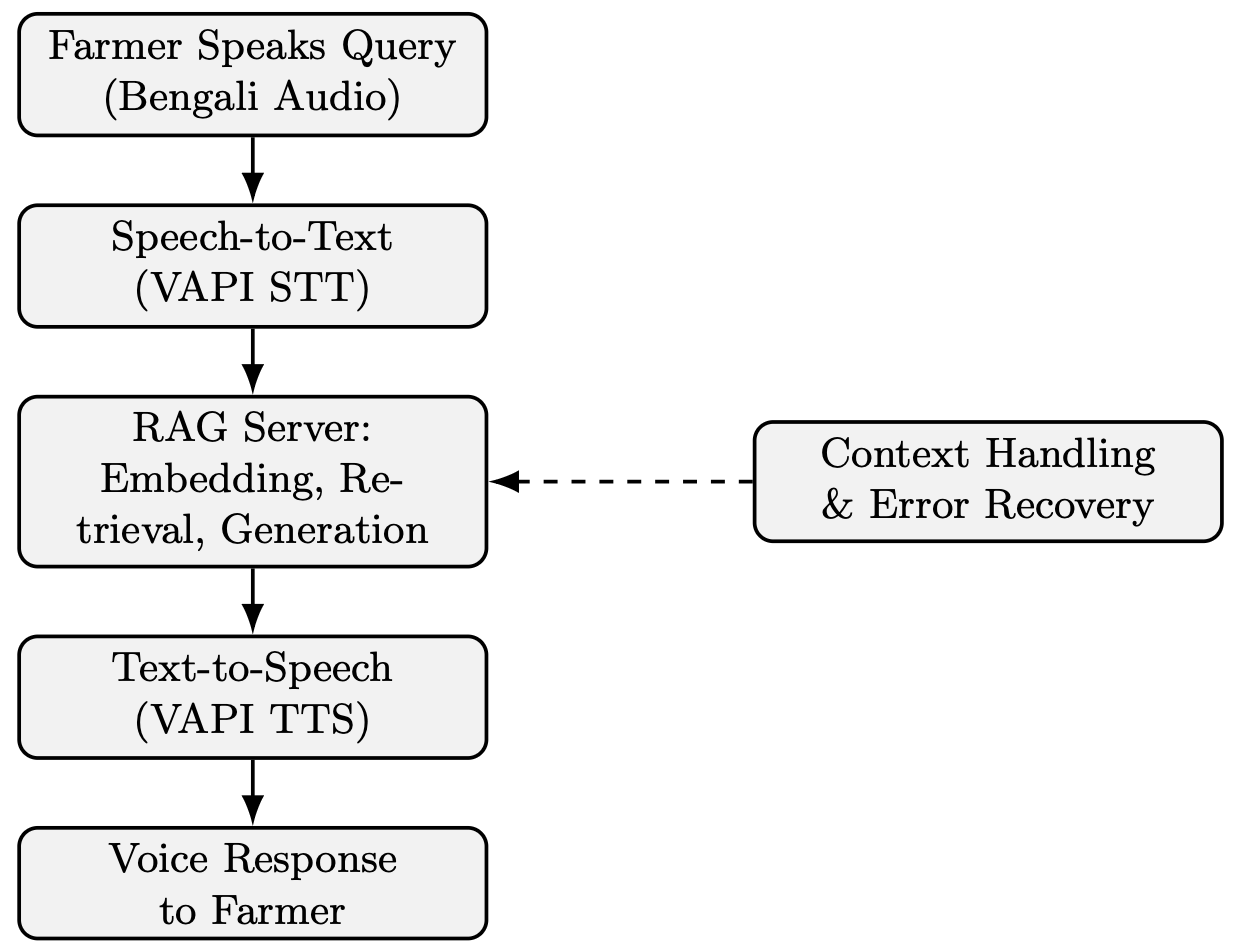}
\caption{Voice Interface Integration Workflow with VAPI}
\label{fig:vapi_pipeline_screenshot}
\end{figure}

VAPI~\cite{vapi2024} enables Bengali speech-to-text (STT) and text-to-speech (TTS) interaction. Operating in a client–server architecture, VAPI handles audio processing while the RAG engine performs retrieval and generation. Figure~\ref{fig:vapi_pipeline_screenshot} shows the end-to-end voice interaction flow integrated with the RAG engine. The system converts spoken queries into text, processes them through the RAG pipeline, and synthesizes responses into natural Bengali speech. Fuzzy matching and contextual dialogue tracking mitigate recognition errors and support multi-turn interaction.

\subsection{Evaluation Methodology}

In the absence of a standardized Bengali agricultural QA benchmark, evaluation was conducted through comparison with Kishan QRS~\cite{kakkar2021kishan}. A representative test set of farmer-style questions was curated across crop diseases, pest control, fertilizers, irrigation, and variety selection. Agricultural experts rated responses as high, moderate, or poor based on accuracy, relevance, and usefulness. For comparative scoring, a 5-point framework assessed relevance, completeness, actionability, contextual richness, and specificity. Semantic similarity (cosine) was additionally computed to quantify retrieval precision and response grounding.

\section{Results and Discussion}

\subsection{Overall System Performance}

\begin{figure}[htbp]
\centerline{\includegraphics[width=0.5\columnwidth]{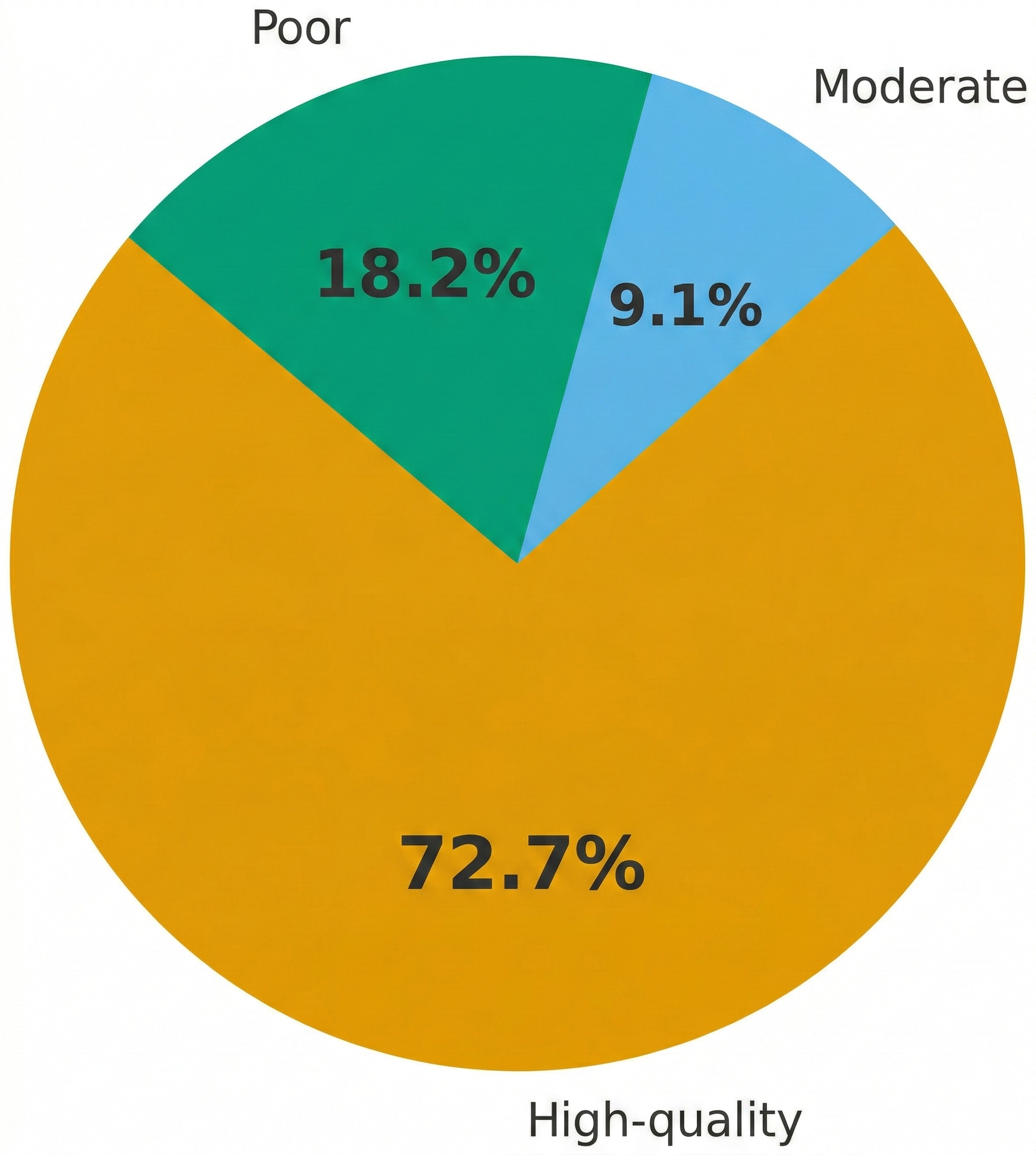}}
\caption{Distribution of response quality categories across test questions, showing strong performance with 72.7\% high-quality responses.}
\label{fig:quality_dist}
\end{figure}

As shown in Figure~\ref{fig:quality_dist}, KrishokBondhu delivered high-quality responses for 72.7\% of test queries, indicating strong reliability in addressing farmers' information needs. Moderate-quality answers (9.1\%) typically arose when relevant context was partially retrieved or when queries required synthesizing dispersed information. Poor responses (18.2\%) were mostly linked to under-represented topics or recent developments not yet present in the knowledge base. Table \ref{tab:sample_responses} presents representative responses from KrishokBondhu across diverse agricultural queries. 

\begin{table*}[htbp]
\centering
\caption{Sample Responses from KrishokBondhu System (Bengali)}
\includegraphics[width=0.8\textwidth]{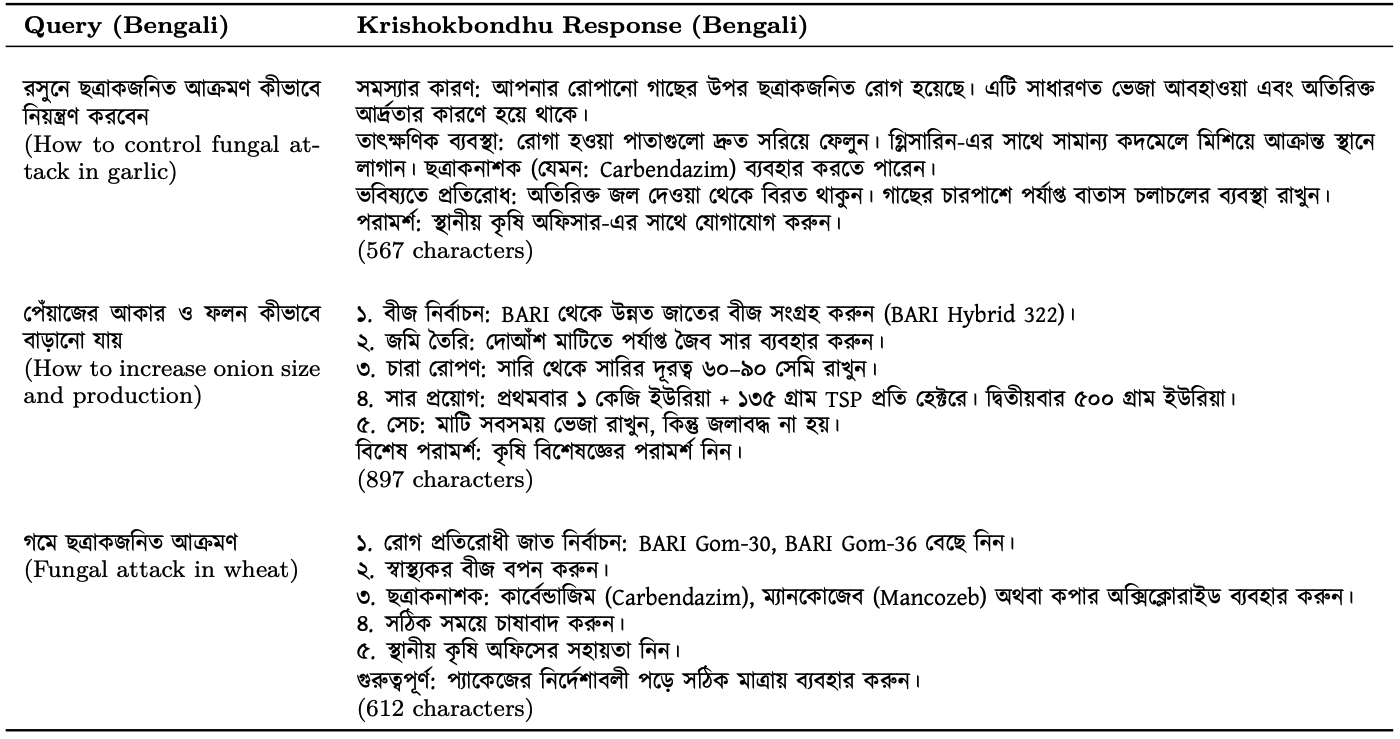}
\label{tab:sample_responses}
\end{table*}

\subsection{Comparative Analysis with Kishan QRS}

To contextualize KrishokBondhu's performance, we conducted a comparative analysis with responses published in the Kishan QRS paper. Table~\ref{tab:sys_comparison} presents matched query-response pairs from both systems, highlighting substantial differences in response characteristics. While Kishan QRS provides concise, technically focused answers averaging 87 characters, KrishokBondhu generates comprehensive responses averaging 692 characters, representing a 7.9-fold increase in detail.

\begin{table}[t]
\footnotesize
\centering
\caption{Comparison between KrishokBondhu and Kishan QRS}
\label{tab:sys_comparison}
\begin{tabular}{p{4.1cm} p{4.1cm}}
\toprule
\textbf{KrishokBondhu (Bengali)} & \textbf{Kishan QRS (English)} \\
\midrule
Retrieval-Augmented Generation (RAG) pipeline using Gemma-3-4B and LanceDB for context-grounded Bengali answers. & Rule-based and deep-learning models trained on historical Kisan Call Centre logs. \\
\addlinespace
Integrates OCR, ASR, and TTS modules for Bengali speech-based interaction. & Text-only interface; accepts English or transliterated Hindi input. \\
\addlinespace
Knowledge base built from BARC’s \textit{Krishi Projukti Hatboi}, DAE manuals, WorldFish repository, and textbooks. & Proprietary dataset from Kisan Call Centre logs; not publicly available or extensible. \\
\addlinespace
Generates detailed responses with explanations, preventive actions, and expert referral guidance. & Produces short prescription-style answers focused on chemical dosage. \\
\addlinespace
Retrieval layer expandable through new document ingestion. & Static dataset; cannot adapt or update post-training. \\
\addlinespace
Supports voice and text I/O for low-literacy farmers via mobile IVR. & Operated by call-centre agents; farmers interact indirectly. \\
\addlinespace
Evaluated on factuality, relevance, and fluency metrics. & Evaluated on text-mapping accuracy only. \\
\bottomrule
\end{tabular}
\end{table}

Quantitative evaluation across five criteria reveals KrishokBondhu's superior performance, as presented in Table~\ref{tab:evaluation_metrics}. The system achieved an overall score of 4.53 out of 5.00, representing a 44.7\% improvement over Kishan QRS's score of 3.13. The most substantial gains appear in contextual richness (4.67 vs 1.00, +367\%) and completeness (4.67 vs 2.33, +100.4\%), reflecting KrishokBondhu's comprehensive approach to agricultural advisory.

\begin{table}[t]
\footnotesize
\centering
\caption{Evaluation Metric Comparison}
\label{tab:evaluation_metrics}
\begin{tabular}{lccc}
\toprule
\textbf{Metric} & \textbf{KrishokBondhu} & \textbf{Kishan QRS} & \textbf{\% Gain} \\
\midrule
Relevance           & 5.00 & 5.00 & +0.0\% \\
Completeness        & 4.67 & 2.33 & +100.4\% \\
Actionability       & 4.33 & 3.33 & +30\% \\
Contextual Info     & 4.67 & 1.00 & +367\% \\
Specific Detail     & 4.00 & 4.00 & +0.0\% \\
\midrule
\textbf{Avg. Score} & \textbf{4.53} & \textbf{3.13} & \textbf{+44.7\%} \\
Resp. Length        & 692 chars & 87 chars & 7.9$\times$ \\
\bottomrule
\end{tabular}
\end{table}

\begin{figure}[htbp]
\centerline{\includegraphics[width=0.9\columnwidth]{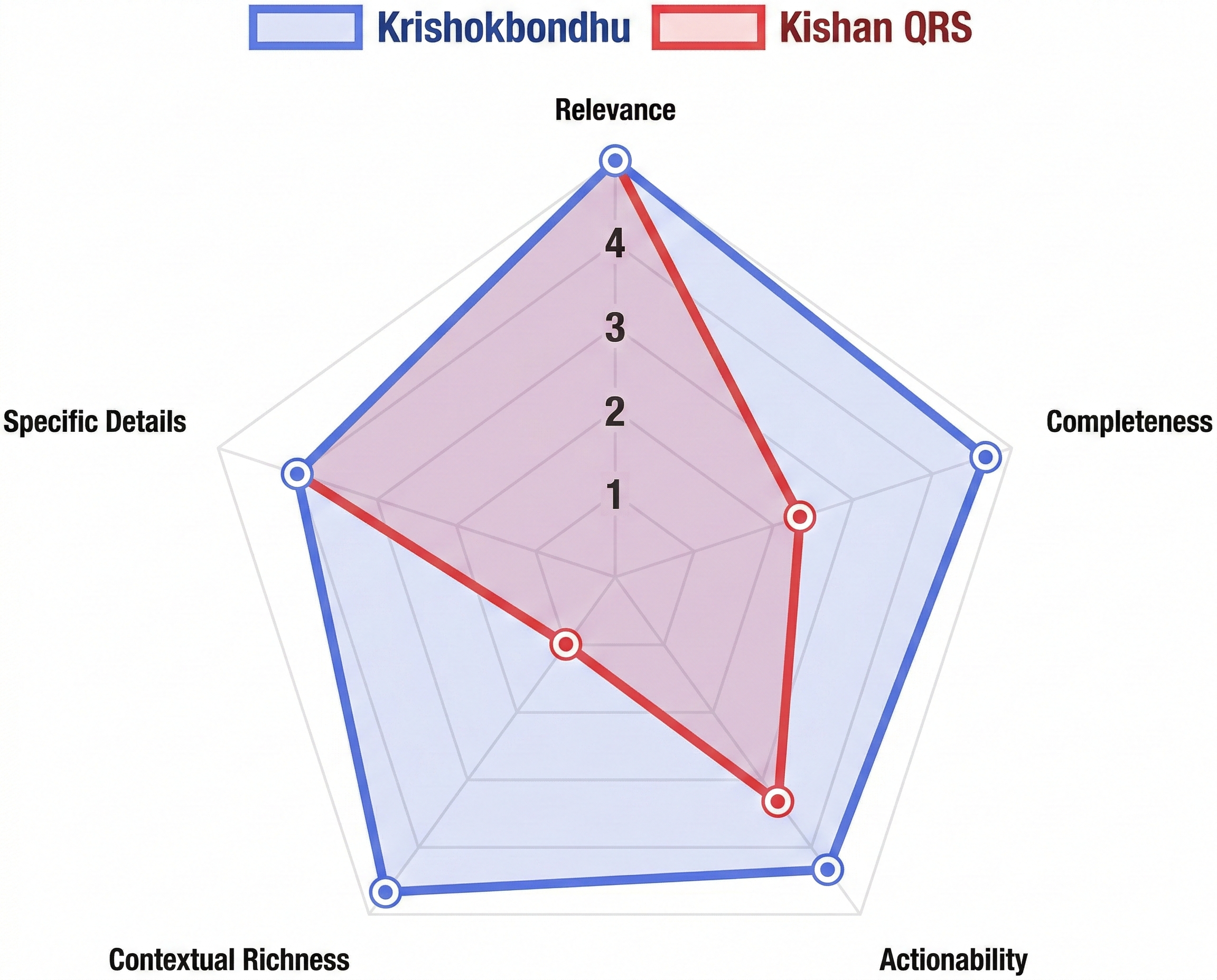}}
\caption{Radar chart comparing KrishokBondhu and Kishan QRS across five evaluation criteria, showing KrishokBondhu's superior performance in contextual richness and completeness.}
\label{fig:radar}
\end{figure}

Figure~\ref{fig:radar} visualizes the comparative performance across evaluation criteria, highlighting KrishokBondhu's particular strengths in providing contextual understanding and comprehensive coverage. Both systems maintain equivalent relevance to queries and similar levels of specific technical details, but KrishokBondhu significantly enhances actionability through step-by-step guidance and completeness through multi-faceted responses.

Information coverage analysis (Table~\ref{tab:coverage_analysis}) reveals distinct differences in response philosophy. KrishokBondhu consistently provides cause explanations (100\% of responses), prevention measures (100\%), and expert referrals (100\%), features largely absent from Kishan QRS responses. This comprehensive approach addresses a critical gap in agricultural extension: farmers often lack understanding of underlying causes, leading to improper implementation of recommendations.

\begin{table}[t]
\centering
\caption{Information Coverage Analysis}
\label{tab:coverage_analysis}
\begin{tabular}{lcc}
\toprule
\textbf{Information Feature} & \textbf{KrishokBondhu} & \textbf{Kishan QRS} \\
\midrule
Cause Explanation         & 3/3 (100\%) & 0/3 (0\%) \\
Immediate Actions         & 3/3 (100\%) & 3/3 (100\%) \\
Prevention Measures       & 3/3 (100\%) & 0/3 (0\%) \\
Specific Dosages          & 2/3 (67\%)  & 3/3 (100\%) \\
Variety Recommendations   & 2/3 (67\%)  & 0/3 (0\%) \\
Expert Referral           & 3/3 (100\%) & 1/3 (33\%) \\
\midrule
\textbf{Average Coverage} & \textbf{83.3\%} & \textbf{38.9\%} \\
\bottomrule
\end{tabular}
\end{table}

\begin{figure}[htbp]
\centerline{\includegraphics[width=\columnwidth]{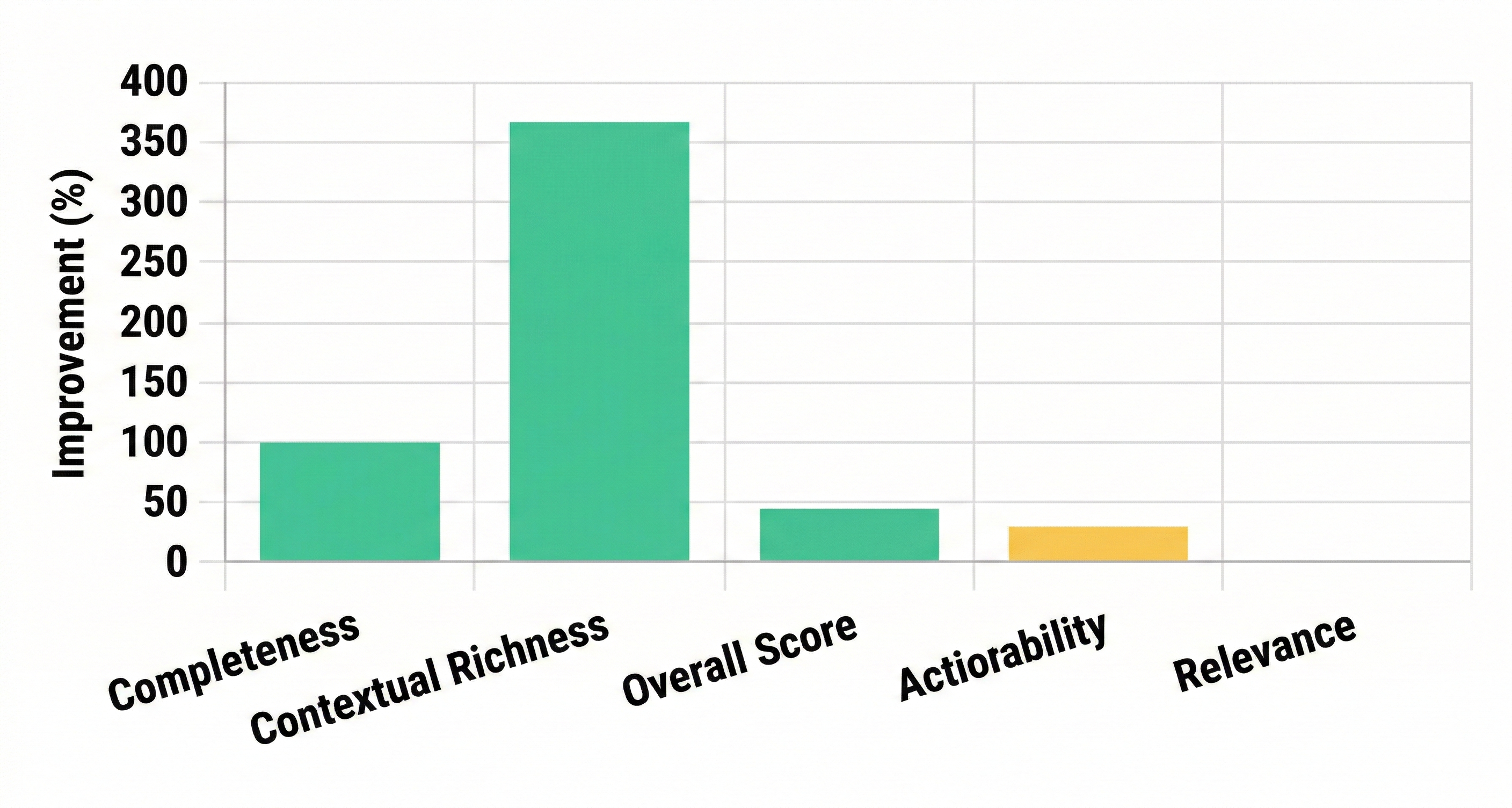}}
\caption{Percentage improvement of KrishokBondhu over Kishan QRS baseline across evaluation criteria, demonstrating substantial gains in contextual richness and completeness.}
\label{fig:improvement}
\end{figure}

Figure~\ref{fig:improvement} illustrates the percentage improvements across evaluation criteria, with contextual richness showing the most dramatic enhancement at 367\%. These improvements validate several design decisions in KrishokBondhu, particularly the RAG architecture's effectiveness in grounding responses in comprehensive agricultural handbooks and the system's optimization for voice-based interaction where detailed explanations enhance farmer understanding.

\subsection{Semantic Similarity and Response Quality Correlation}

\begin{figure}[htbp]
\centerline{\includegraphics[width=0.9\columnwidth]{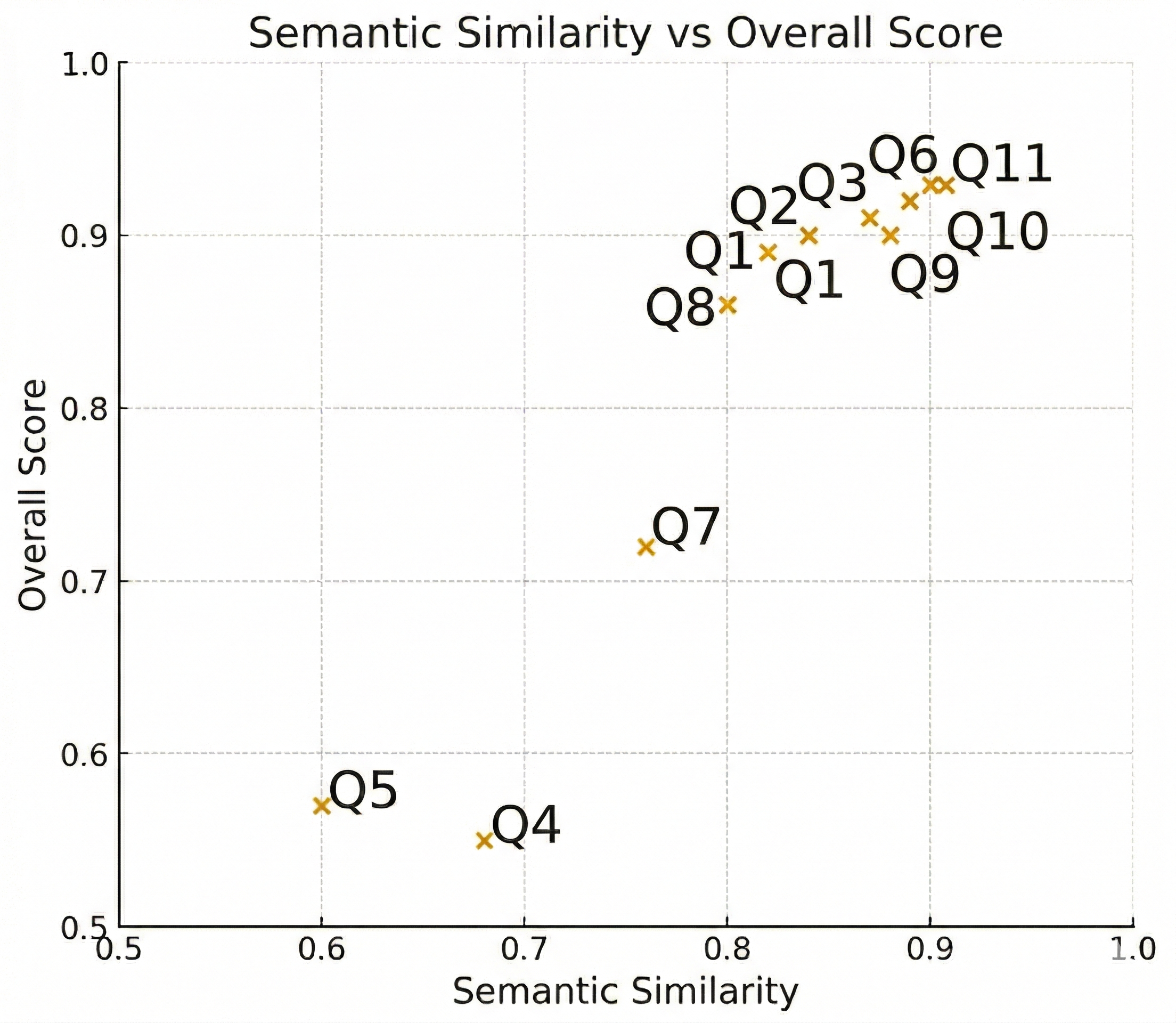}}
\caption{Relationship between semantic similarity and overall quality scores, showing positive correlation between retrieval quality and response assessment.}
\label{fig:similarity}
\end{figure}

Figure~\ref{fig:similarity} illustrates a clear positive correlation between semantic similarity and response quality. Most high-quality responses had similarity scores above 0.85, indicating strong alignment between user queries and retrieved content. This suggests that effective retrieval is a strong predictor of answer quality.

Interestingly, a few lower-similarity cases still produced good answers—often when information was scattered across multiple segments or required broader reasoning. On the other hand, some high-similarity matches led to weaker responses, underscoring that retrieval precision, not similarity alone, is critical for consistent quality.

\section{Conclusion}

KrishokBondhu demonstrates the feasibility of building a voice-enabled, retrieval-augmented agricultural advisory system grounded in authoritative local documentation. By combining document digitization, semantic retrieval, large language models, and Bengali speech interaction, the system provides farmers with accessible, context-aware guidance aligned with national agricultural practices. The integration of voice input and output further lowers literacy barriers and supports inclusive knowledge dissemination in rural settings.

While the current implementation shows strong potential, several limitations remain. The evaluation is constrained by the absence of standardized Bengali agricultural QA benchmarks and relies on expert-based assessment. In addition, the knowledge base requires continuous updating to reflect newly released crop varieties, emerging diseases, and evolving cultivation practices. 

Future work will focus on expanding document coverage, incorporating structured knowledge representations alongside vector retrieval, integrating real-time contextual signals such as weather or market data, and exploring domain-specific model adaptation. Overall, this work highlights the promise of retrieval-augmented NLP systems in strengthening agricultural knowledge access in low-resource environments and advancing AI-driven extension services in Bangladesh.

\bibliographystyle{IEEEtran}
\bibliography{references}

\end{document}